\documentclass{article}
\usepackage{spconf,amsmath,graphicx}
\usepackage{subfigure}
\usepackage{xcolor}

\title{Calibrationless Parallel MRI using Model based deep learning (C-MODL) }
\name{Aniket Pramanik, Hemant Aggarwal, Mathews Jacob\thanks{This work is supported by NIH 1R01EB019961-01A1.}}
\address{The University of Iowa, Iowa City, USA}
\begin{document}
%
\maketitle
\begin{abstract}
We introduce a fast model based deep learning approach for calibrationless parallel MRI reconstruction. The proposed scheme is a non-linear generalization of structured low rank (SLR) methods that self learn linear annihilation filters from the same subject. It pre-learns non-linear annihilation relations in the Fourier domain from exemplar data. The pre-learning strategy significantly reduces the computational complexity, making the proposed scheme three orders of magnitude faster than SLR schemes. The proposed framework also allows the use of a complementary spatial domain prior; the hybrid regularization scheme offers improved performance over calibrated image domain MoDL approach. The calibrationless strategy minimizes potential mismatches between calibration data and the main scan, while eliminating the need for a fully sampled calibration region. 
\end{abstract}
\vspace{-1em}
\section{Introduction}
\label{sec:intro}
Modern MRI schemes often rely on the spatial diversity of the coil sensitivities to recover the images from undersampled k-space measurements. Pre-calibrated, auto-calibrated, and calibrationless methods have been introduced for parallel MRI recovery. The celebrated SENSE approach is a pre-calibrated scheme, where the coil sensitivities are estimated from separate calibration scans \cite{pruessmann1999sense}. Several priors including total variation and deep learned networks \cite{aggarwal2018modl} have been introduced to regularize image recovery in the SENSE approach. A challenge with calibration based methods is the potential for motion artifacts, which results from mismatches between the calibration and main scans. Autocalibrating methods \cite{griswold2002generalized,lustig2010spirit} overcome this challenge by learning the coil sensitivities from fully sampled center of k-space (autocalibrating signal), while allowing undersampling at higher frequencies. Recently, deep learning methods \cite{akccakaya2019scan} that learn a neural network from the auto-calibration region have also been introduced to improve the image quality at high acceleration factors. Despite these advances, the need to acquire fully-sampled auto-calibration regions often restricts the achievable acceleration and the amount of higher k-space samples that can be acquired in a realistic scan time. To minimize these trade-offs, several researchers have introduced calibrationless parallel structured low-rank (PSLR) matrix completion approaches \cite{shin2014calibrationless,haldar2013low}. These methods lift the Fourier coefficients of a multi-channel image into a Toeplitz/Hankel matrix, which is low-rank due to linear annihilation relations within the multi-channel k-space data. The PSLR schemes exploit this low-rank property to interpolate the missing matrix entries, and hence fill in the missing k-space entries. These methods offer good performance at a high computational complexity. Specifically, the estimation of the signal/null subspaces of the Hankel matrix from the measured data by the iterative PSLR schemes is computationally expensive.

The main focus of this work is to introduce a calibrationless parallel MRI strategy using deep learning, which is computationally more efficient than PSLR algorithms. Similar to PSLR methods, the proposed scheme works in k-space to exploit annihilation relations in the multi-channel k-space data. The key distinction of the proposed scheme with PSLR methods is that the annihilation relations are pre-learned from exemplar data. We note that an iterative least squares algorithm to solve the PSLR problem involves the alternation between a data consistency enforcement step and a projection step, which constrains the k-space data to the estimated signal subspace. In PSLR methods, the subspace is self-learned from the data itself in an iterative fashion, which contributes to bulk of the computation time. By contrast, the use of a pre-learned CNN instead of the self-learned linear projection significantly reduces the computational complexity. Unlike direct inversion based deep-learning methods \cite{han2019k, el2019calibrationless}, we rely on a model based approach similar to \cite{aggarwal2018modl}; the proposed k-space network architecture has multiple repetitions of denoising and data consistency enforcing blocks with shared parameters across iterations. The network is unrolled and trained in an end-to-end fashion. Our experiments show that the proposed k-space network provides reconstructions that are comparable in quality to calibrationless PSLR schemes, while offering several fold reduction in the run time. \\
In this work, we additionally rely on spatial domain deep learning priors, which learns the structure of patches in images,  along with the above k-space annihilation relations. The resulting hybrid unfolded network uses both k-space and image space denoising blocks in an unrolled fashion. Our experiments show that the hybrid approach provides improved image quality compared to the k-space alone network. This work is a generalization of our previous work \cite{pramanik2019off}, which focussed on the recovery of single channel MRI data.
\vspace{-1em}
\section{Background}
\subsection{Problem setup}
Parallel MRI aims to recover the image $\rho$ from the Fourier coefficients of the  coil sensitivity weighted images
\begin{eqnarray}\
\label{coilimages}
\rho_{i}(\mathbf{r})=\rho (\mathbf{r})~s_i (\mathbf{r}), i =1, .. N,
\end{eqnarray}
where  $s_i(\mathbf r)$ is the coil sensitivity of the $i^{\rm th}$ coil and $\mathbf r$ denotes spatial domain coordinates. The acquisition can be modeled as $\mathbf{b}_i = \mathcal{A}(\widehat{\boldsymbol{\rho}_i}) + n_i, \forall i=1,2 \ldots N$, 
where $\widehat{\boldsymbol{\rho}_i}$ are Fourier coefficients of $i^{th}$ image, $N$ is the total number of coils, $n_i$ is the corresponding zero mean Gaussian noise, and $\mathcal A$ is an undersampling operator.
\begin{figure}[t!]
\begin{center}
\subfigure[Linear residual denoising block used in PSLR ]{\includegraphics[width=0.5\textwidth,keepaspectratio=true,trim={5cm 7cm 5cm 7cm},clip]{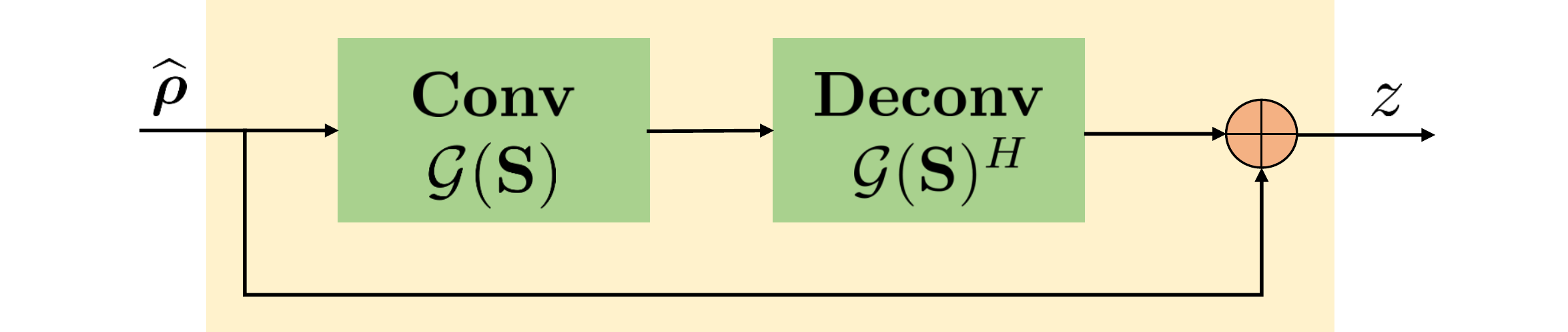}}
\subfigure[Residual CNN denoiser]{\includegraphics[width=0.5\textwidth,keepaspectratio=true,trim={0cm 6cm 0 2.5cm},clip]{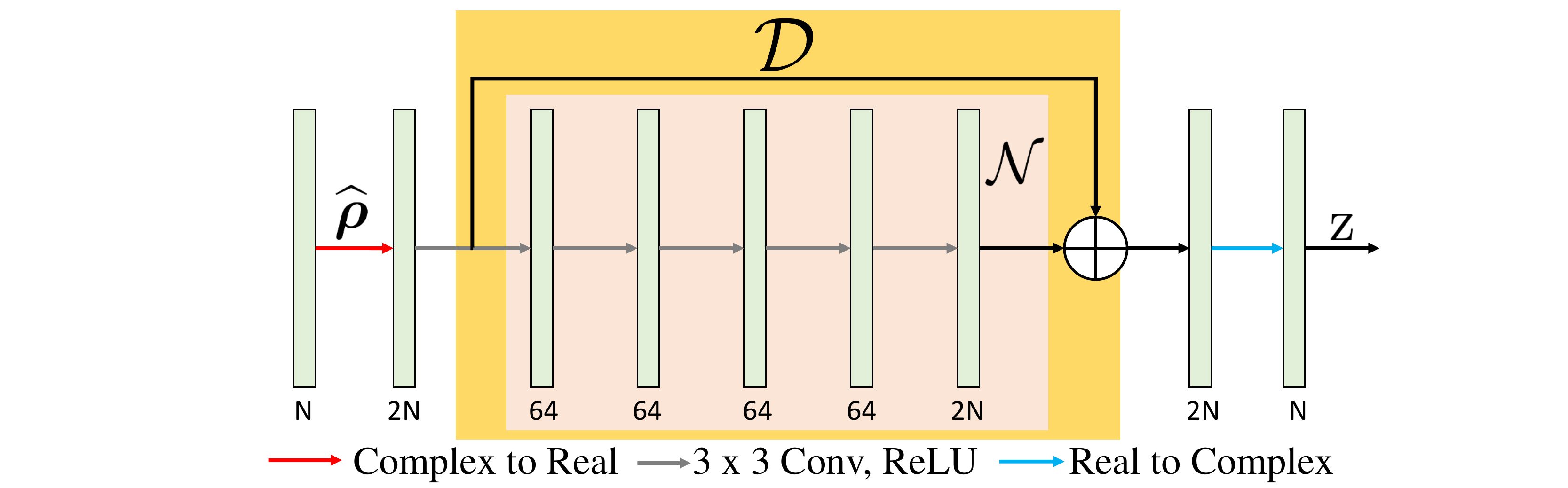}\label{fig:cnn}}
\subfigure[Proposed K-space network]{\includegraphics[width=0.45\textwidth,keepaspectratio=true,trim={4.5cm 6cm 4.2cm 5cm},clip]{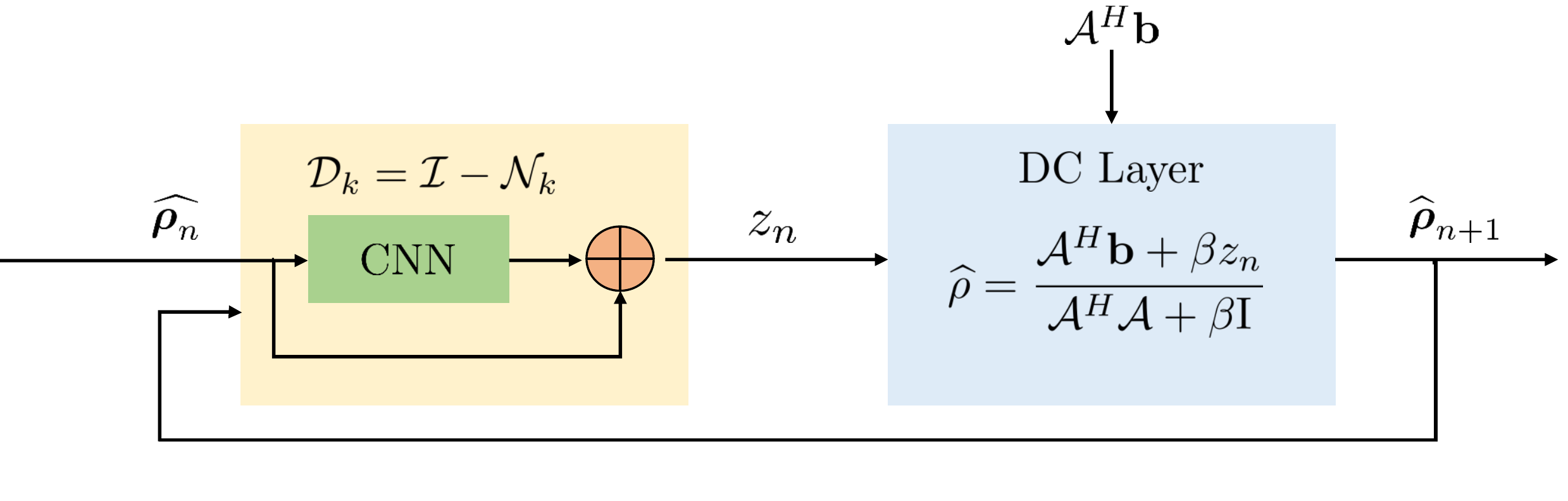}\label{fig:ksp_net}}
\subfigure[Proposed Hybrid network]{\includegraphics[width=0.45\textwidth,keepaspectratio=true,trim={4cm 6cm 3.5cm 5cm},clip]{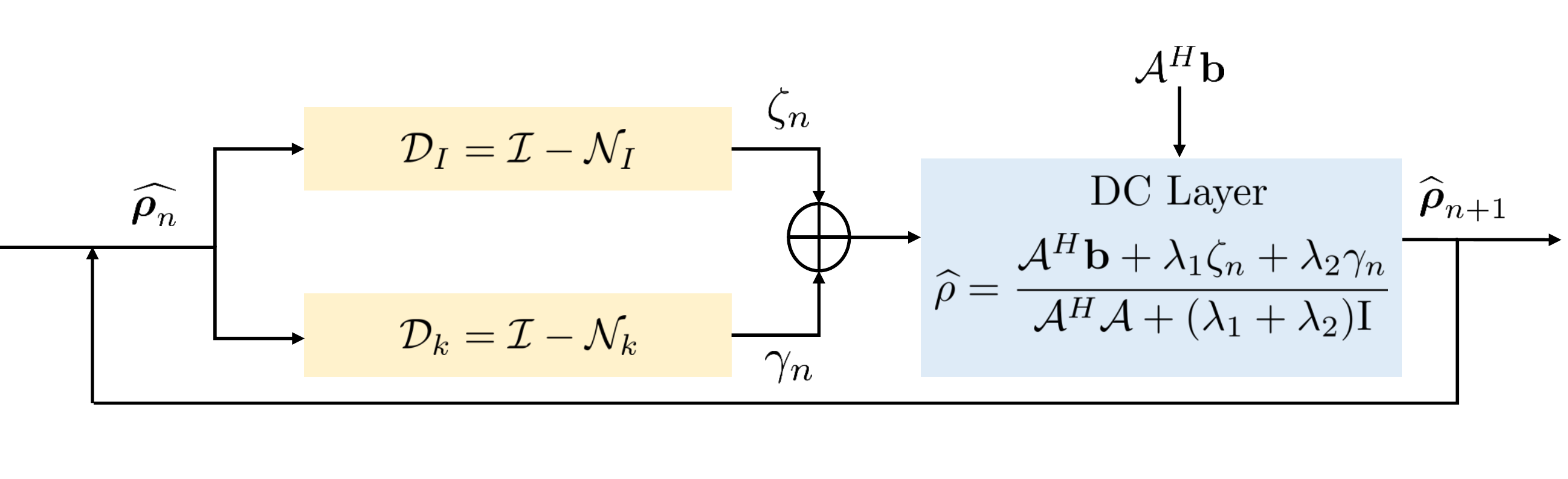}\label{fig:hyb_net}}
\end{center}
\vspace{-1.2em} 
\caption{Outline of the recursive learning architectures. (a) PSLR residual denoising block, replaced by the CNN in (b); (b) Proposed residual denoising block; (c) K-space network, $\mathcal{D}_k$ is a residual CNN from (b); (d) Hybrid network with $\mathcal{D}_k$ and $\mathcal{D}_I$ being identical residual CNNs for k-space and image domain learning respectively.
}
\vspace{-1.2em} 
\end{figure}
\vspace{-1em}
\subsection{Parallel Structured Low Rank (PSLR) recovery}
The images in \eqref{coilimages} satisfy a spatial domain annihilation relation $\rho_{i}(\mathbf{r})s_{j}(\mathbf{r})-\rho_{j}(\mathbf{r})s_{i}(\mathbf{r})=0, \forall \mathbf{r}$  \cite{morrison2007multichannel}, which translates to convolution in k-space
\begin{equation}
\label{conv_pi}
\widehat{\rho_{i}}[\mathbf{k}]\ast\widehat{s_{j}}[\mathbf{k}]-\widehat{\rho_{j}}[\mathbf{k}]\ast\widehat{s_{i}}[\mathbf{k}]=0, \forall \mathbf{k}.
\end{equation}
Here, $\widehat{s_{j}}[\mathbf{k}]$ is Fourier domain representation of $s_{j}(\mathbf{r})$ and $\mathbf k$ denotes k-space coordinates. 
The above k-space annihilation relation can be rewritten as a block Hankel matrix multiplication $
\mathbf{H} (\widehat{\rho_{i}})\cdot\widehat{s_{j}}-\mathbf{H} (\widehat{\rho_{j}})\cdot\widehat{s_{i}}=0$, where $\mathbf{H}(\widehat{\rho_{i}})\cdot\widehat{s_{j}}$ corresponds to  2-D convolution between $\widehat{\rho_{i}}$ and $\widehat{s_{j}}$. Combining the annihilation relation for every pair of coil images, we obtain $
\mathcal{T}(\widehat{\rho})\cdot \mathbf{S} = 0$,
where $\mathcal{T}(\widehat{\rho})=\left[\begin{array}{cccc}\mathbf{H} (\widehat{\rho_{1}})
& \mathbf{H} (\widehat{\rho_{2}}) & \ldots & \mathbf{H} (\widehat{\rho_{N}})\end{array}\right]$ and the columns of $\mathbf S$ correspond to the concatenation of Fourier coefficients $\widehat s_i$.
The Hankel matrix $\mathcal{T}(\widehat{\rho})$ has a huge set of null space vectors in $\mathbf{S}$ which implies it is low-rank. The recovery of $\widehat{\boldsymbol \rho}$ is posed as
\begin{equation}
\label{slr}
\arg \min_{\widehat{\boldsymbol \rho}} \|\mathcal{A}(\widehat{\boldsymbol \rho})-\mathbf b\|_2^2 + \lambda \|\mathcal{T}(\widehat{\boldsymbol \rho})\|_\ast
\end{equation}
where $\|\cdot\|_{\ast}$ is nuclear norm penalty that encourages $\mathcal{T}(\widehat{\boldsymbol \rho})$ to have a large null-space. 
\vspace{-1em}
\section{Model based Calibrationless PMRI }
The structured low-rank optimization problem in \eqref{slr} can be solved using an iterative reweighted least-squares algorithm which majorizes the nuclear norm with  a weighted Frobenius norm $\|\mathcal{T}(\widehat{\boldsymbol \rho})\|_\ast \leq \|\mathcal{T}(\widehat{\boldsymbol \rho})\mathbf S\|_F^2 = \|\mathcal{G}(\mathbf S ) \widehat {\boldsymbol \rho}\|_F^2 $. Here, the columns of $\mathbf S$ are estimated from the current iterate as 
\begin{equation}\label{nullspaceupdate}
\mathbf S = [\mathcal{T}^{\ast}(\mathcal{T}(\widehat{\boldsymbol \rho}))+\epsilon \mathbf I]^{-1/4}
\end{equation}
Since convolution is a commutative operation, we can rewrite $\mathcal{T}(\widehat{\boldsymbol \rho})~ \mathbf S$ as $\mathcal{G}(\mathbf S)~ \widehat{\boldsymbol \rho}$, where $\mathcal G(\mathbf S)$ is an appropriately sized block Hankel matrix composed of the columns of $\mathbf S$. This allows us to rewrite $ \|\mathcal{T}(\widehat{\boldsymbol \rho})\mathbf S\|_F^2 = \|\mathcal{G}(\mathbf S ) \widehat {\boldsymbol \rho}\|_F^2$. The above relations results in the iterative strategy that alternates between \eqref{nullspaceupdate} and
\begin{equation}
\label{irls}
\widehat{\boldsymbol \rho} = \arg \min_{\widehat{\boldsymbol \rho}} \left \| \mathcal{A}(\widehat{\boldsymbol \rho})-\mathbf{b}\right \|_2^2 + \lambda \left \| \mathcal{G}(\mathbf S)~ \widehat{\boldsymbol \rho}\right\|_F^2
\end{equation}
We note that the term $\mathcal{G}(\mathbf S ) \widehat {\boldsymbol \rho}$ correspond to the convolution of the multichannel input $\widehat{\boldsymbol{\rho}}$ with a multichannel filterbank, whose filters are the columns of $\mathbf S$. 
\vspace{-1em}
\subsection{Residual Conv-Deconv structure of PSLR}
Using an auxiliary variable $\mathbf z$ in \eqref{irls} and using a penalty term to impose the constraint $ \mathbf z= \widehat{\boldsymbol \rho}$, we rewrite \eqref{irls} as
\begin{equation}
\arg \min_{\widehat{\boldsymbol \rho}} \|\mathcal{A}(\widehat{\boldsymbol \rho})-\mathbf b\|_2^2 + \lambda \|\mathcal{G}(\mathbf S)\mathbf z\|_F^2+ \beta \| \widehat{\boldsymbol \rho} - \mathbf z\|_2^2.
\end{equation}
Note that the above expression is equivalent to \eqref{irls} as $\beta \rightarrow \infty$. The IRLS solution alternates between the following sub-problems:
\begin{eqnarray}\label{analytic_sub}
\mathbf {\widehat{\boldsymbol \rho}}_{n+1} &=& \arg \min_{\widehat{\boldsymbol \rho}}  \|\mathcal A \widehat{\boldsymbol \rho} - \mathbf b\|^2_2 + \beta \|\widehat{\boldsymbol \rho} -  \mathbf z_{n}\|^2_2 \\
\label{denoiser_sub}
\mathbf z_{n+1} &=& \arg \min_{\mathbf z} \beta \|\widehat{\boldsymbol \rho}_{n+1} - \mathbf z\|^2_2+ \lambda \|\mathcal G(\mathbf S_n)\mathbf z\|_F^2
\end{eqnarray}
We note that \eqref{analytic_sub} can be solved analytically. Solving \eqref{denoiser_sub} we get $ \mathbf z = \left[\mathbf I ~+~\frac{\lambda}{\beta}~\mathcal G(\mathbf S_n)^H \mathcal G(\mathbf S_n) \right]^{-1} \widehat{\boldsymbol \rho}$. Assuming $\lambda <<\beta$ and using matrix inversion lemma:
\begin{equation}
\label{denoiser_gen}
\mathbf z_{n+1} \approx \underbrace{\left[\mathbf I ~-~\frac{\lambda}{\beta}~\mathcal G(\mathbf S_n)^H \mathcal G(\mathbf S_n) \right]}_{\mathcal L_n} \widehat{\boldsymbol \rho}_{n+1}
\end{equation}

\begin{table}[t!]
\fontsize{8}{16}
\selectfont
\centering
\renewcommand{\arraystretch}{0.7}
\begin{tabular}{|c|ccc|cc|}
\hline
\multicolumn{4}{|c|}{Brain} & \multicolumn{2}{|c|}{Knee} \\ \hline
Methods & 6x & 10x & Run time & 4x & Run time \\ \hline 
PSLR &21.02 &18.12 &367s &24.26 &932s\\
K-space UNET &19.58 &17.28 &0.21s &27.01 &0.85s\\
\textbf{Proposed k-space} &\textbf{21.58} &\textbf{18.71} &\textbf{0.05s} &\textbf{27.87} &\textbf{0.19s}\\
MoDL &23.30 &21.63 &0.25s &29.77 &1.32s\\
\textbf{Proposed hybrid} &\textbf{24.34}  &\textbf{22.20} &\textbf{0.11s} &\textbf{30.57} &\textbf{0.42s}\\ \hline
\end{tabular}
\caption{Quantitative comparison of PSLR, MoDL, proposed and UNET reconstructions with run times. The average SNR is reported in dB and run time in seconds.}
\label{tab:comp} 
\vspace{-1em}
\end{table}
Note that $\mathcal G(\mathbf S)^H$ involves multichannel convolution with flipped conjugate filters, and hence can be viewed as a \emph{deconvolution} multichannel filter-bank. Thus, $\mathcal L_n$ can be viewed as a residual linear block consisting of a convolution and deconvolution filter banks in series. ${\mathcal L}_n$  essentially projects $\widehat{\boldsymbol \rho}_{n+1}$ to the signal subspace, thus \emph{denoising} the current iterate. The multi-channel filterbank $\mathbf S_n$ is updated at each step of the iterative algorithm; the joint update of the variables is associated with high computational complexity. Note that auto-calibration strategies often estimate $\mathbf S$ from calibration data \cite{griswold2002generalized,lustig2010spirit,uecker2014espirit}; in this case, one can use a constant $\mathcal L$ that does not change with iterations, thus quite significantly reducing the computational complexity.  
\vspace{-1em}
\subsection{Model based deep learning}
We propose to use a fixed non-linear CNN $\mathcal D_k$ instead of the linear filter-bank $\mathcal L_n$, which varies from iteration to iteration to reduce the computational complexity of the PSLR algorithm. The subscript of $\mathcal D_k$ indicates that the denoiser works in k-space. We conjecture that the generalizability of CNN blocks facilitates the pre-learning of the multi-channel relations from exemplar data. The proposed CNN $\mathcal D$ is shown in Fig \ref{fig:cnn}, where the $\mathcal N$ block consists of 5 layers of convolution with ReLU non-linearity embedded between them. The $\mathcal D$ block consists of $\mathcal N$, followed by adder similar to \eqref{denoiser_gen}.
  
Similar to MoDL \cite{aggarwal2018modl}, our network has a recursive structure, alternating between $\mathcal D$ and data-consistency enforcing blocks (denoted by DC) for a fixed number of iterations $K$. The unrolled network is trained end-to-end with shared network parameters across iterations. A notable difference with MoDL is that the proposed network learns parameters in k-space. The k-space network is shown in Fig \ref{fig:ksp_net}.
Motivated by the ability of image domain priors used in \cite{aggarwal2018modl}, we propose to use an additional image domain prior that exploits the redundancy of patches in the spatial domain, along with the k-space CNN. The proposed hybrid consists of two 5-layer CNNs (proposed in \ref{fig:cnn}) denoted as $\mathcal D_k$ and $\mathcal D_I$, working in parallel as shown in Fig \ref{fig:hyb_net}; the $\mathcal D_k$ and $\mathcal D_I$ blocks learn k-space and spatial domain relations respectively. 
\vspace{-1em}
\section{Experiments and Results}
\begin{figure*}[t!]
\begin{center}
\includegraphics[width=\textwidth,keepaspectratio=true,trim={2cm 8.5cm 2.2cm 8.5cm},clip]{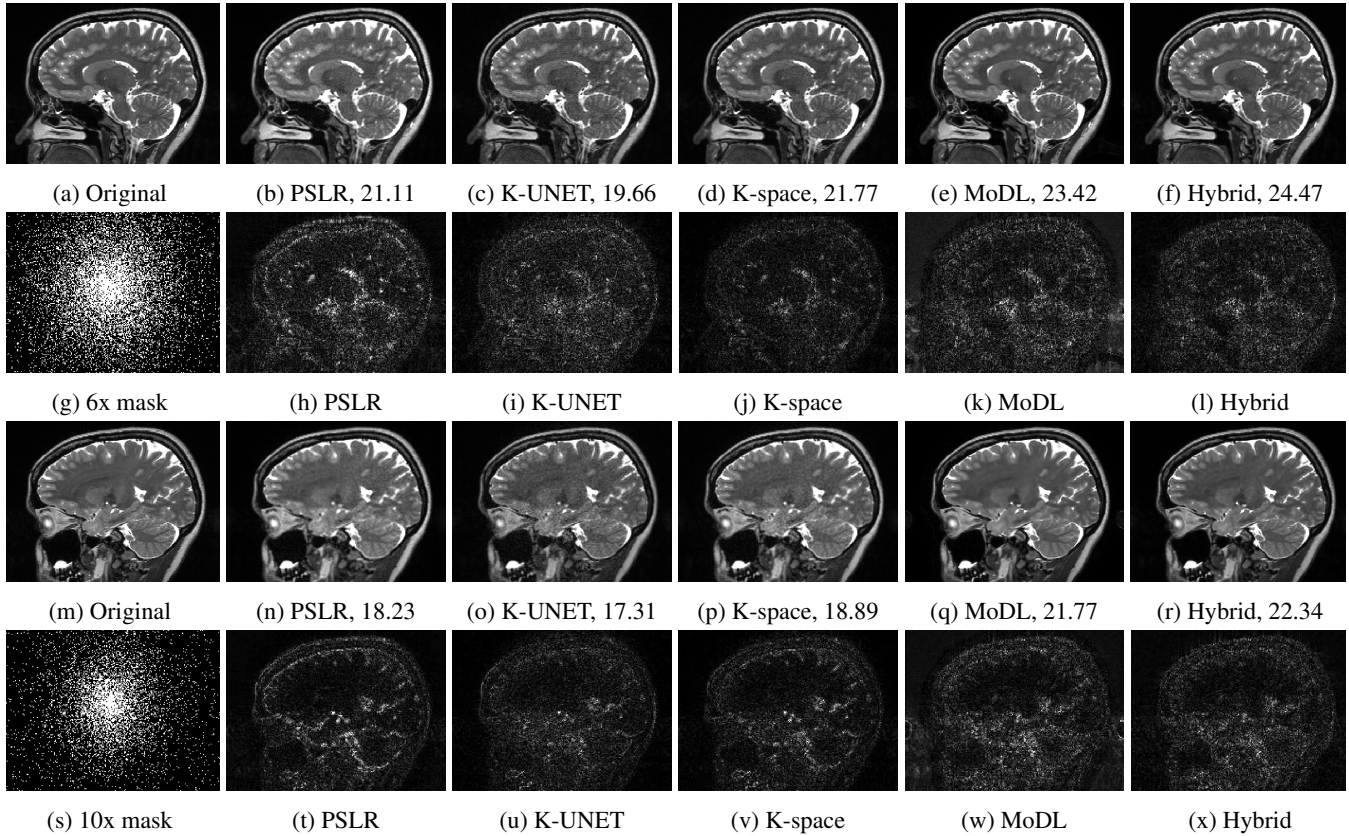}
\end{center}
\vspace{-0.75cm}
\caption{Reconstruction results of 6x and 10x accelerated brain data. K-UNET stands for k-space UNET \cite{han2019k}, while K-space is the proposed scheme with k-space network alone and Hybrid is the proposed scheme with both k-space and image space networks. MoDL is a calibration-based scheme that uses the coil-sensitivities, which are pre-estimated; (a)-(l) and (m)-(x) show results with corresponding absolute error images for 6x and 10x undersampling respectively.}\vspace{-1em}
\label{fig:recon}
\end{figure*}
We conduct experiments on brain data collected from five subjects scanned at the University of Iowa Hospitals. A 3D T2 CUBE sequence was applied for Cartesian readout using a 12-channel head coil. Four subjects (90 x 4 = 360 slices) were used for training and the remaining one for testing. The experiments were conducted using a 2D non-uniform cartesian variable density undersampling mask with different acceleration factors; the readout direction was orthogonal to the slices. We also ran simulations on publicly available knee data \cite{hammernik2018learning} with 4x cartesian undersampling along phase encodes.
 
We compare the proposed k-space alone (K-space) and hybrid (Hybrid) algorithms against PSLR \cite{shin2014calibrationless}, k-space UNET (K-UNET) \cite{han2019k} and MoDL. Both knee and brain reconstruction results with run times are recorded in Table \ref{tab:comp}. Note that MoDL is a calibrated deep learning approach, which additionally uses coil sensitivities to recover images in spatial domain; it uses more prior information, compared to all the other methods. K-UNET is a direct deep learning approach consisting of a 20 layer UNET without a DC step. Reconstructions of brain for 6x and 10x undersampling are shown in Fig \ref{fig:recon}.

The K-space network performs better than both K-UNET and PSLR, but lags behind MoDL that uses additional coil sensitivity information and spatial regularization. We note that the hybrid approach, which includes an image domain prior together with the k-space network improves the performance, thus outperforming the calibration-based MoDL scheme. The calibration-free strategy is desirable since it eliminates the vulnerability to potential mismatches between calibration and main scans.
\vspace{-1em}
\section{Conclusion}
We propose a calibrationless model based deep learning for Parallel MRI reconstruction. It is a non-linear extension of recent calibrationless structured low-rank methods for Parallel MRI that rely on annihilation relations in Fourier domain due to complementary information among coil images. The proposed method learns non-linear annihilation relations in the Fourier domain from exemplar data. The non-linearity helps to generalize learned annihilation relations over images unseen during training unlike structured low-rank methods which self learn linear annihilation relations from the input image every time. The proposed scheme is significantly faster than the SLR methods. We add an image domain prior to propose a hybrid network consisting of both k-space and spatial domain CNNs. The hybrid network out-performs the calibrated MoDL framework and other state-of-the-art calibrationless approaches.         
\vspace{-1em}
\bibliographystyle{IEEEbib}

\bibliography{refs_ap}

\end{document}